# Leveraging Internal Representations of Model for Magnetic Image Classification

Adarsh N L, Dr. Arun P V, Dr. Alok Porwal, Dr. Malcolm Aranha

*Abstract*—Data generated from edge devices hold the potential for training intelligent autonomous systems across multiple domains. Various machine learning approaches have emerged to address privacy issues and leverage distributed data, but security concerns persist due to privacy-sensitive storage of data shards in different locations. In this paper, we propose a potentially novel paradigm for training machine learning models, particularly in scenarios where only a single magnetic image and its corresponding labelled image are available. We exploit the potential of Deep Learning to create limited but informative samples for training a Machine Learning model. By leveraging the power of deep learning's internal representations, we aim to overcome data scarcity and achieve meaningful results in a resource-efficient manner. This approach represents a promising direction for training machine learning models with minimal data.

## I. INTRODUCTION

MAGNETIC image classification has a wide range of applications and plays a crucial role in various fields, including medical diagnosis, materials science, and geophysics.

Traditionally, experts have relied on manual analysis and domain-specific knowledge to classify these images, but this process can be time-consuming and subjective. With the advent of machine learning, feature extraction techniques have become crucial in optimizing downstream tasks. In [?], Kumar et al. discuss how these techniques can be used to improve the accuracy and efficiency of magnetic image classification.

In [2], F. I. Hasib et al. focus on magnetic configuration classification using deep neural network (DNN) approaches. Their work provides a valuable resource for researchers in magnetic materials and image data classification. Al-Saffar et al.'s work [3] explores the power of convolutional neural networks (CNNs) in image classification. Their findings provide insights into research challenges and future directions.

Rekha et al. [4] emphasized the importance of Earth observations and satellite remote sensing in Indian agriculture, providing valuable insights for precision agriculture. In their study, Ball et al. [5] explored the possibilities of deep learning (DL) in remote sensing (RS), shedding light on the challenges and opportunities, and demonstrating DL's efficacy in handling hyperspectral and multispectral data applications.

A significant improvement in large-scale image classification using the Deep Multi-Task Learning algorithm was presented by [6], Kuang et al. Their approach efficiently categorized millions of images into thousands or tens of thousands of object classes, showcasing its potential for practical applications. On a different front, Mou et al. [7] tackled the limitations of vector-based machine learning algorithms in hyperspectral image classification by proposing an innovative deep recurrent neural network (RNN) model.

A novel approach for remote sensing image change detection was introduced by Chen et al. [8], who unveiled the bitemporal image transformer (BIT). This method outperformed conventional techniques, delivering superior results with significantly lower computational costs and parameters. Additionally, in [9, 10], SVM-based methods were explored. The study in [9] conducted a comprehensive review of support vector machines (SVMs) in the context of remote sensing, highlighting their compelling advantages for analyzing airborne and satellite-derived imagery. In [10], the authors proposed an SVM-based region-growing algorithm for extracting urban areas from DMSP-OLS and SPOT VGT data, providing a semi-automatic approach for extracting urban extents. Moreover, they presented an alternative implementation technique for SVM, offering an effective solution for small-sized training datasets in hyperspectral remote sensing data.

The research work [12, 13] focused on the role of ontologies in interpreting remote sensing images and bridging the gap between expert expectations and the contribution of ontologies in remote sensing research. Kwenda et al.'s survey [12] provided insights into machine-learning methods for forest image analysis and classification, while Arvor et al.'s work [13] emphasized the importance of knowledge-driven approaches in remote sensing.

Proposing a template matching-based object recognition technique, [16] introduced a method for locating objects in images with unknown geometric parameters. In the study by G. Simone et al. [18], image fusion techniques for remote sensing applications were showcased, covering diverse methods such as synthetic aperture radar (SAR) interferometry, multisensor and multitemporal image fusion, and SAR image fusion. The paper discussed the application and advantages of these techniques in the context of remote sensing endeavours. In [14], a comprehensive review of wetland remote sensing was presented, encompassing various sensors and applications, rendering it a valuable resource for researchers studying wetland ecosystems. Furthermore, Smith and Pain [15] extensively explored the utilization of remotely sensed imagery in geomorphology since the early availability of Landsat data, emphasizing the transformative impact of new technologies.

Hojat Shirmard et al.'s review [19] covered machine learning in processing remote sensing data for mineral exploration, emphasizing the potential of advanced methods like deep learning. Xiong et al.'s study [20] applied big data analytics and deep learning algorithms for mapping mineral prospectivity, demonstrating promising results. The papers [21, 22] focused on GIS-based mineral prospectivity mapping using machine learning methods, with machine learning models outperforming traditional methods. Luke Smith et al. [24] discussed the application of deep learning super-resolution architectures for enhancing grid resolution in magnetic surveys. Additionally, the papers [25, 26] addressed the challenges in interpreting complex aeromagnetic data and detecting first breaks in seismic refraction data using machine learning methods, demonstrating their potential in geological interpretations.

In this introduction, we presented an overview of the revolutionary impact of ML and DL on magnetic image classification and remote sensing. However, these approaches pose challenges such as the need for substantial labeled data and significant computational resources. Obtaining labelled data for large-scale landscapes in remote sensing can be time-consuming and costly. Additionally, the opacity of deep learning models hinders their interpretability, which can be critical in applications requiring explainability.

To address these issues, this research paper proceeds with a detailed discussion of the data used and the process of leveraging internal representations of a network to train an ML model for magnetic image classification (Section II). Subsequently, we delve into the results of this approach in Section III, aiming to shed light on these challenges and provide valuable insights for future advancements in magnetic image classification and remote sensing. The overall workflow is picturized in Fig.1.

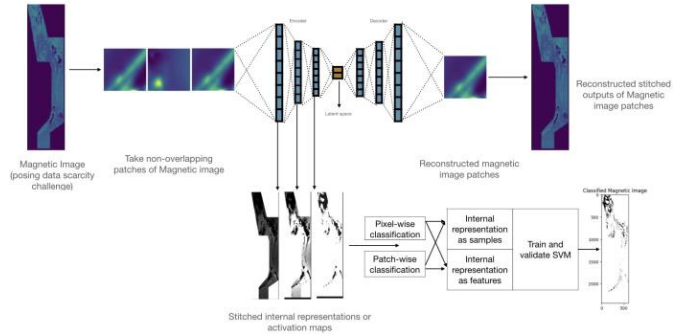

Fig. 1. The overall workflow of the research

## II. Data and Methodology

### A. Data

The data used in this research is displayed in Fig.2. It is a single-band magnetic image of a terrain (Left) and on the right is a corresponding pixel-wise manually labelled image. The shape of the image is (2434, 607). The unique labels in the labeled image are 0,1,2 corresponding to non-deposit, deposit, and unknown or unexplored respectively. There are 14 deposit pixels and 17 non-deposit pixels, and the remaining pixels are unknown in the image. Since the number of labelled pixels is very low and the image is a single-band magnetic image, traditional data augmentation methods will not be suitable here, and training a model directly becomes challenging. Therefore, there is a clear need for alternative approaches to effectively train a model for accurate classification.

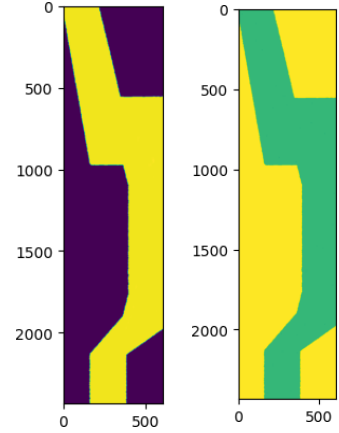

Fig. 2. The Magnetic Image data (Left) and the corresponding pixel-wise manually labelled image data (Right)

### B. Methodology

As there is a data scarcity problem imposed, the methodology is designed to tackle it effectively. Given the limited number of labelled pixels in the single-band magnetic image, a different approach is necessary. To address this challenge, we adopted a patch-based strategy to train an autoencoder. Specifically, we extracted patches of size 50x50 around each pixel. With this approach, we now have a total of 1477438 patches of size 50x50. Few sample patches are displayed in Fig. 3. By utilizing these localized patches, the autoencoder is able to learn meaningful representations from the magnetic image data, despite the scarcity of labelled samples. This approach aims to enhance the model's ability to discern distinct patterns and features within the image, thereby improving the classification accuracy in the presence of limited training data.

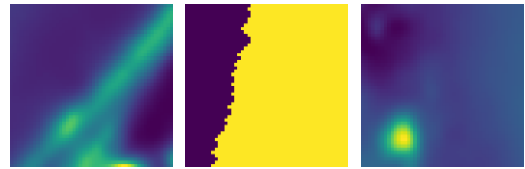

Fig. 3. Random 50x50 patches of the Magnetic Image data

The architecture of the autoencoder used for training is displayed in Fig. 4. Once the autoencoder is trained, we try to leverage the internal representations of the magnetic image to achieve a more informative and discriminative feature space for classification. To accomplish this, we first reshape the magnetic image into a format suitable for extracting layer-wise representations from the trained autoencoder. This reshaping process involves breaking down the image into patches of size 50x50, forming a data matrix.

Next, we utilize the trained autoencoder as a feature extractor by creating an activation model. This activation model takes the reshaped patches as inputs and returns the intermediate activations (layer-wise representations) of the model up to the third layer (until the encoder layer). These activations capture meaningful feature representations that reflect the underlying structures and patterns in the magnetic image data.

The resulting layer-wise representations are then reshaped and stitched back together to reconstruct the original magnetic image. This stitching process combines the individual patches with overlap to ensure that the internal representations from neighbouring patches are seamlessly integrated. The overlap helps maintain continuity and reduce artifacts in the final reconstructed image. Fig.5. displays the reconstructed stitched image and an intermediate stitched image.

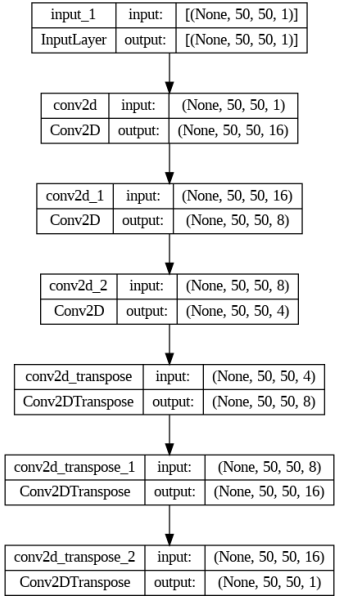

Fig. 4. The architecture of the autoencoder used for training

The stitched images based on the internal representations from the autoencoder are stored in a list. Each element of the list corresponds to one of the filters of intermediate layers of the autoencoder. By considering these images at different layers, we aim to capture hierarchical and increasingly complex features that can lead to better discrimination between deposit and non-deposit regions during classification.

Overall, this process allows us to transform the magnetic image into a feature-rich representation using the autoencoder's internal layers, thereby enhancing the model's ability to classify accurately, even with limited labelled data.

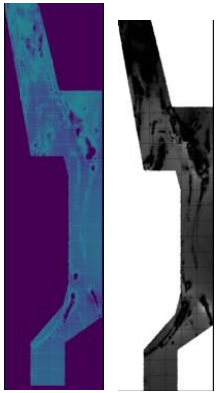

Fig. 5. Reconstructed stitched image of the autoencoder (Left). Intermediate layer stitched image (Right).

## III. EXPERIMENTS AND RESULTS

With these stitched internal representations or activation maps of the magnetic image, we experiment with two approaches for training a machine-learning model. Firstly, we consider the layer-wise stitched image as samples and then we also experiment with considering features for classifying the magnetic image data in Fig.2. both pixel-wise as well as patch-wise. The results of these experiments are displayed in the table.

TABLE I
ACCURACY AND F1 SCORE OF THE SVM MODELS

|  | Pixel-wise classification | | Patch-wise classification | |
| --- | --- | --- | --- | --- |
|  | Features | Samples | Features | Samples |
| Accuracy | 80 % | 68.93 % | 71.4 % | 67 % |
| F1 score | 0.6 | 0.6 | 0.65 | 0.66 |

### A. Pixel-wise classification

In the pixel-wise classification experiments, the objective was to classify individual pixels within the magnetic images as deposit or non-deposit areas. The internal representations of the autoencoder were utilized for training the machine learning model.

*1) Internal representations as samples:* For this approach, we randomly selected 24-pixel positions from the label image, ensuring a balanced representation with 12 pixels from deposit areas and 12 from non-deposit areas. The pixel values at these sample positions were extracted from the stitched images and used as samples for training the SVM classifier. K-fold cross-validation with k=5 was employed to evaluate the model's accuracy across different folds. The accuracy scores for each fold were recorded, and the mean accuracy across all folds was calculated. Subsequently, the classifier was trained on the entire training data, and its overall accuracy was tested. The results indicated an overall accuracy of approximately 68.93 %.

*2) Internal representations as features:* In this sub-subsection, we again used the SVM classifier and focused on the internal representations or activation maps of the magnetic image as features for training the machine learning model. We randomly selected 24 sample positions from the label image and obtained pixel values at the same location across all stitched images. These pixel values were then used to construct feature vectors for each sample. Similar to the previous approach, we employed k-fold cross-validation with k=5 to assess the model's accuracy across different folds. The mean accuracy across all folds was computed, and the classifier was trained on the entire training data. This approach yielded promising results, achieving an accuracy of approximately 80 %.

### B. Patch-wise classification

Moving on to patch-wise classification, we aimed to classify patches extracted from stitched magnetic images as either deposit or non-deposit areas. Once again, the internal representations of the autoencoder were employed for training the SVM model.

*1) Internal representations as samples:* In this sub-subsection, we explored the patch-wise classification of magnetic images using internal representations as samples. A specified number of sample positions were randomly selected for both deposit and non-deposit classes, representing the center coordinates of the patches to be extracted. The patches were then extracted around these sample positions from each stitched image and resized to a consistent size of 10x10 pixels. The SVM classifier was trained using the extracted patches, achieving an overall accuracy of approximately 66.6 %

*2) Internal representations as features:* Lastly, the patch-wise classification was examined using internal representations as features. Similar to the previous approach, we randomly selected sample positions and extracted patches around these positions from each stitched image. The pixel values within the patches were then used to construct feature vectors. The SVM classifier was trained using these feature vectors, resulting in an overall accuracy of approximately 71.4 %.

Overall, the experiments demonstrated the effectiveness of leveraging internal representations of magnetic images for pixel-wise and patch-wise classifications. The classified magnetic image is displayed in Fig. The combination of autoencoder-generated internal representations and SVM classifiers showcased promising results in distinguishing between deposit and non-deposit areas, offering potential applications in geological interpretations and mineral mapping.

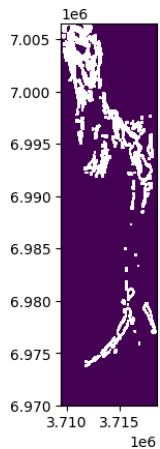

Fig. 6. Pixel-wsie classified Magnetic image using the SVM model (Activation maps as features).

## IV. CONCLUSION

In this paper, we propose a new way to train machine learning models when there is not a lot of data available. Our approach uses the internal representations of deep learning models to overcome the challenges of limited data availability, data distribution, and privacy concerns issues.

We have shown that our approach is effective in enhancing training efficiency and improving model performance. Our work opens up new avenues for training machine learning models with limited data and offers a promising solution for optimizing model training while preserving privacy.

In the future, we plan to further refine our techniques for leveraging internal representations, develop robust strategies to handle security concerns and explore the applicability of our approach in various domains and datasets.